\documentclass[letterpaper]{article} 
\usepackage{aaai2027}  
\usepackage[hyphens]{url}  
\usepackage{graphicx} 
\usepackage{natbib}  
\usepackage{caption} 
\usepackage{algorithm}
\usepackage{algorithmic}
\usepackage{amsmath}
\usepackage{amssymb}
\usepackage{multirow}

\usepackage{newfloat}
\usepackage{listings}
\DeclareCaptionStyle{ruled}{labelfont=normalfont,labelsep=colon,strut=off} 
\floatstyle{ruled}
\newfloat{listing}{tb}{lst}{}
\floatname{listing}{Listing}
\usepackage{bm}
\usepackage{booktabs}

\title{A2DINOv3: Rethinking Multi-Modal Object Detection via Socialized Collaboration}
\author {
    Jiekang Feng{\rm 1}\equalcontrib,
    Zhihe Fan{\rm 2}\equalcontrib,
    Yunqi Zhu{\rm 3},
    Xinjie Yao{\rm 4}\corresponding,
    Yueying Zhang{\rm 1},
    Yike Gao{\rm 1},
    Ranxin Li{\rm 1},
    Guanzuo Chen{\rm 1}
    Pengfei Zhu{\rm 5}
}
\affiliations {
    \textsuperscript{\rm 1}School of Artificial Intelligence, Tianjin University\\
    \textsuperscript{\rm 2}School of Sports Training, Tianjin University of Sport\\
    \textsuperscript{\rm 3}School of Computer Science and Engineering, University of New South Wales\\
    \textsuperscript{\rm 4}Faculty of Information Engineering and Automation, Kunming University of Science and Technology\\
    \textsuperscript{\rm 5} School of Automation, Southeast University
}

\begin{document}

\maketitle

\begin{abstract}
Multi-modal object detection is essential for robust scene understanding in challenging conditions, including low-light and adverse environments. Recent vision foundation models (e.g., DINOv3) have exhibited strong representation capabilities, yet adapting them to multi-modal scenarios remains challenging. Existing dense cross-modal fusion strategies often force heterogeneous modalities to interact indiscriminately, which may introduce redundant information and disrupt the valuable pre-trained representations. To address this issue, we revisit multi-modal fusion from the perspective of socialized learning and propose adapter to DINOv3 (A2DINOv3), a multi-expert collaboration framework with a Socialized Collaboration Protocol (SCP). Specifically, RGB and infrared branches are modeled as heterogeneous experts that independently preserve their specialized knowledge while exchanging complementary information through selective and constrained interactions. This design mitigates harmful cross-modal interference and prevents degradation of pre-trained priors during adaptation. Furthermore, a zero-initialization strategy is introduced to gradually activate cross-modal collaboration, enabling a smooth transition from modality-specific learning to cooperative representation learning. Extensive experiments on four multi-modal benchmarks, including aerial detection (GAIIC), autonomous driving (FLIR), low-light surveillance (LLVIP), and diverse real-world scenarios (M3FD), demonstrate that A2DINOv3 consistently achieves state-of-the-art performance in multi-modal object detection.
\end{abstract}

\section{Introduction}

\label{sec:intro}

DINOv3~\cite{DINOv3} and its predecessors~\cite{dino}, as vision foundation models pre-trained on large-scale data, have enabled new paradigms for object detection when integrated with detection architectures like DETR~\cite{DETR}. By learning transferable visual priors from massive unlabeled data, these foundation models exhibit strong generalization ability across diverse downstream tasks. Nevertheless, their capability is bounded by the information availability of the input modality. In real-world scenarios with severe visual degradation, including nighttime environments, intense illumination changes, and atmospheric interference, single-modal sensors often fail to provide sufficient and reliable cues, motivating the exploration of multi-modal perception paradigms that can exploit complementary sensory information.

\begin{figure}[!t]
\vspace{-10pt}
\centering
\includegraphics[width=\columnwidth,height=0.48\textheight,keepaspectratio]{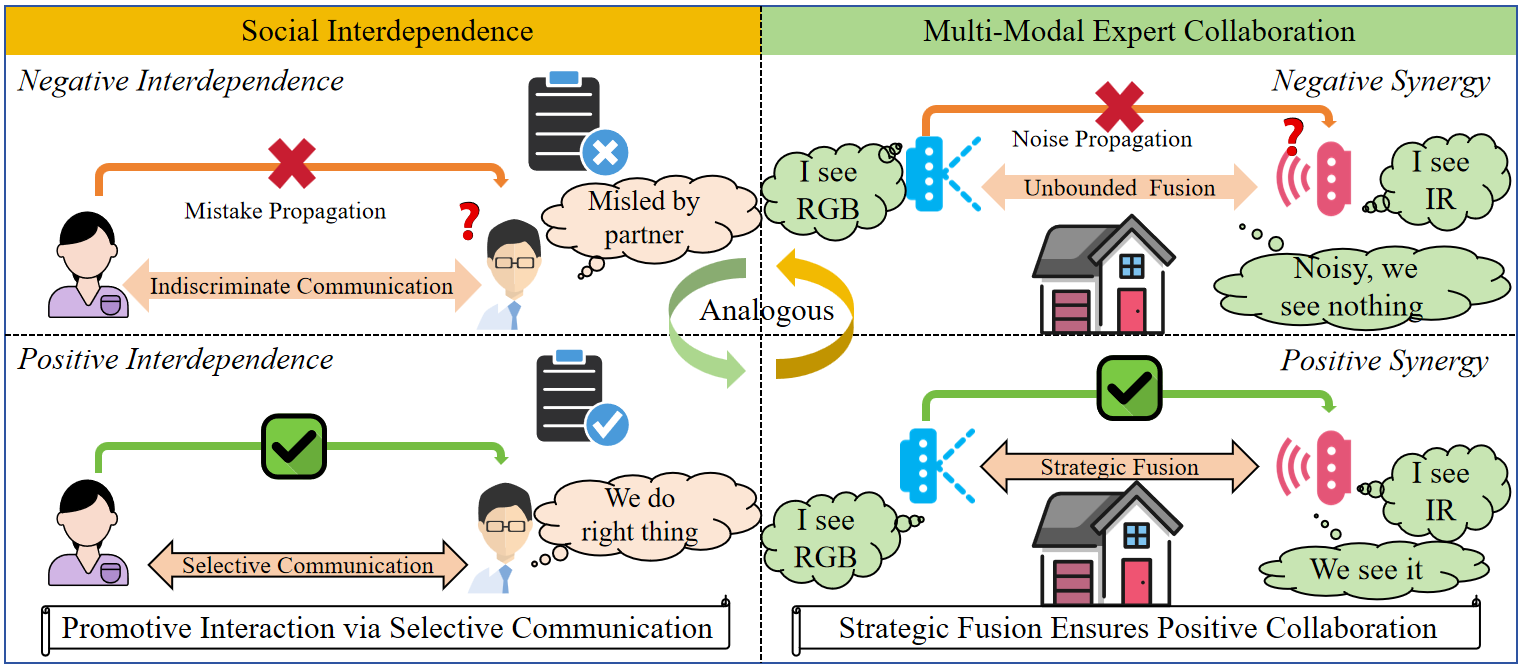}
\vspace{-10pt}
\caption{Social interdependence in human society versus socialized interdependence in multi-modal object detection.}
\label{fig:teaser}
\vspace{-17pt}
\end{figure}

Multi-modal perception offers a potential solution to overcome the information bottleneck of single-sensor detectors by enabling complementary evidence aggregation across heterogeneous modalities~\cite{muti1,muti2}. RGB and IR sensors exhibit distinct yet complementary characteristics: RGB captures fine-grained appearance information when illumination is sufficient, whereas IR provides robust structural and thermal cues under degraded visibility conditions. Despite this complementarity, directly coupling these modalities with vision foundation models remains non-trivial, as indiscriminate cross-modal interactions may introduce irrelevant information and, more critically, cause harmful cross-modal interference that degrades the valuable pre-trained priors. Therefore, the key challenge lies not only in fusing multi-modal features, but in determining how foundation models should selectively collaborate with heterogeneous modalities during adaptation.


Although vision foundation models provide powerful transferable representations, how to effectively adapt them to heterogeneous multi-modal scenarios remains an open problem. Existing RGB-IR detectors~\cite{ICAFusion} typically rely on task-specific convolutional architectures, while parameter-efficient tuning methods~\cite{lora,vpt} and recent DETR-based approaches~\cite{deim} mainly focus on single-modal adaptation. They lack a mechanism to regulate cross-modal collaboration while preserving the intrinsic knowledge encoded by foundation models. As shown in Fig.~\ref{fig:teaser}, naive adaptation suffers from two coupled issues: (1) Fusion Degradation, where excessive cross-modal communication introduces unreliable signals and triggers negative transfer under degraded conditions; and (2) Feature Distortion, where aggressive optimization alters the pre-trained representation space and compromises the foundation model's original visual priors. These challenges reveal that multi-modal adaptation is not merely a feature fusion problem, but an interaction regulation problem.


To address these challenges, we draw inspiration from social interdependence theory in social psychology~\cite{co_a,co_b} and propose adapter to DINOv3 (A2DINOv3). The theory suggests that group outcomes are determined not merely by whether individuals interact, but by how their interdependence is structured. Unregulated interaction can produce negative interdependence, where unreliable members interfere with collective performance. This principle closely parallels multi-modal detection: dense and indiscriminate communication allows noise from a degraded modality to propagate into the other branch together with useful cues, leading to cross-modal interference and fusion degradation. By contrast, positive interdependence preserves the autonomy of individual members while enabling selective cooperation toward a shared objective. From this perspective, effective multi-modal fusion should not enforce unrestricted feature sharing, but should allow each modality to retain its specialized representation and exchange only information that contributes to joint detection.


Specifically, A2DINOv3 treats the RGB and IR branches as heterogeneous experts and coordinates them through bounded communication. We introduce bidirectional Socialized Collaboration Protocol (SCP), whose narrow bottlenecks compress modality-specific features into compact interaction cues, limiting low-level noise while preserving task-relevant information. Zero-initialized up-projections keep the communication paths inactive at the start of training, allowing each branch to retain its pre-trained representation. As optimization proceeds, task-driven gradients progressively activate cross-modal interaction, enabling the model to learn when information exchange is beneficial and how strongly it should affect the receiving branch. The refined multi-level evidence is finally integrated via mean fusion to yield a unified prediction. In essence, A2DINOv3 shifts multimodal fusion from unrestricted feature mixing to regulated information exchange, facilitating complementary collaboration without compromising modality-specific pre-trained priors. Our main contributions are summarized as follows:


\begin{itemize}
    \item We introduce a socialized collaboration for multi-modal foundation model adaptation, viewing heterogeneous modalities as independent experts that require regulated information exchange rather than exhaustive fusion.
    \item We propose A2DINOv3, a framework equipped with a socialized collaboration protocol and implicit curriculum learning to progressively activate cross-modal interaction while preserving pre-trained visual priors.
    \item Extensive experiments on four multi-modal benchmarks demonstrate that A2DINOv3 achieves state-of-the-art performance and consistently improves robustness under diverse challenging scenarios.
\end{itemize}

\section{Related Works}
\label{sec:related_work}
\subsection{Multi-Modal Fusion}

\textbf{Multi-Modal Fusion} in visible-infrared object detection is a prominent paradigm in computer vision that aims to explore cross-modal feature complementarity, where feature-level fusion has become the dominant approach. Existing methods can be broadly categorized into two main groups: \textbf{(1) Spatial-focused architectures} mainly rely on dynamically weighted fusion based on cues such as illumination~\cite{36,59,63}, and subsequently introduce various dense attention-based interaction modules into cross-modal settings~\cite{12}. \textbf{(2) Frequency-focused architectures} attempt to extract or fuse complementary information directly in the frequency domain. In addition to the aforementioned approaches, other methods exist outside these categories that specifically target challenging perception scenarios. For instance, QFDet~\cite{56} designs a label assignment strategy for tiny objects by combining prior and posterior knowledge, while COXNet ~\cite{27} captures targets in complex environments through multi-scale cross-modal alignment, and IM-CMDet~\cite{25} strengthens signals via differential fusion and feature reconstruction.

Existing methods rely on dense cross-modal interaction, which can propagate unreliable signals and distort pre-trained representations. The challenge is to exploit complementarity without compromising modality-specific competence. A2DINOv3 addresses this trade-off through SCP, which restrict information exchange to suppress noise while preserving useful cross-modal cues.

\subsection{Adapter-Based Transfer Learning}
 
\textbf{Adapter-based transfer learning} is a parameter-efficient paradigm that adapts pretrained models by introducing lightweight trainable modules while freezing most backbone parameters. Existing methods can be broadly categorized into two groups: (1) \textbf{Architecture-focused adapters} modify the placement or structure of adapter modules. AdaptFormer~\cite{adapterformer} inserts parallel adapters into ViT MLP sub-blocks, while RepAdapter~\cite{luo2023towards} reparameterizes adapters into the backbone to eliminate additional inference latency. (2) \textbf{Optimization-focused adapters} regulate how newly introduced modules participate in training. LLaMA-Adapter~\cite{zhang2023llamaadapter} employs zero-initialized scalar gates to progressively activate adapter outputs, whereas AdapterTune~\cite{khazem2026adaptertune} zero-initializes the up-projection matrix to produce an initially inactive residual path. Recent works explore efficient adapter designs, including teacher-guided adaptation~\cite{zhang2026teaadapter} and lightweight quantization-aware adapters~\cite{mohammadi2026fixing}.

Existing adapters primarily support efficient single-stream adaptation rather than heterogeneous cross-modal communication. A2DINOv3 repurposes low-rank adapters as bidirectional SCP pathways, progressively activated through zero-initialized up-projections to enable selective transfer while preserving modality-specific representations.

\section{Methodology}
\label{sec:method}

We present A2DINOv3, a multi-modal detector that treats RGB–IR fusion as structured collaboration between heterogeneous experts rather than unrestricted mixing. Its design follows three principles: modality-specific competence preservation, capacity-constrained communication, and progressive dependency emergence. These are instantiated via a shared dual-stream backbone, a Socialized Collaboration Protocol (SCP), and zero-initialized pathways, with features aggregated through fusion and fed to the detection head.

\subsection{Motivation and Overview}
\label{sec:overview}

\begin{figure*}[!t]
    \centering
    \includegraphics[width=0.99\textwidth]{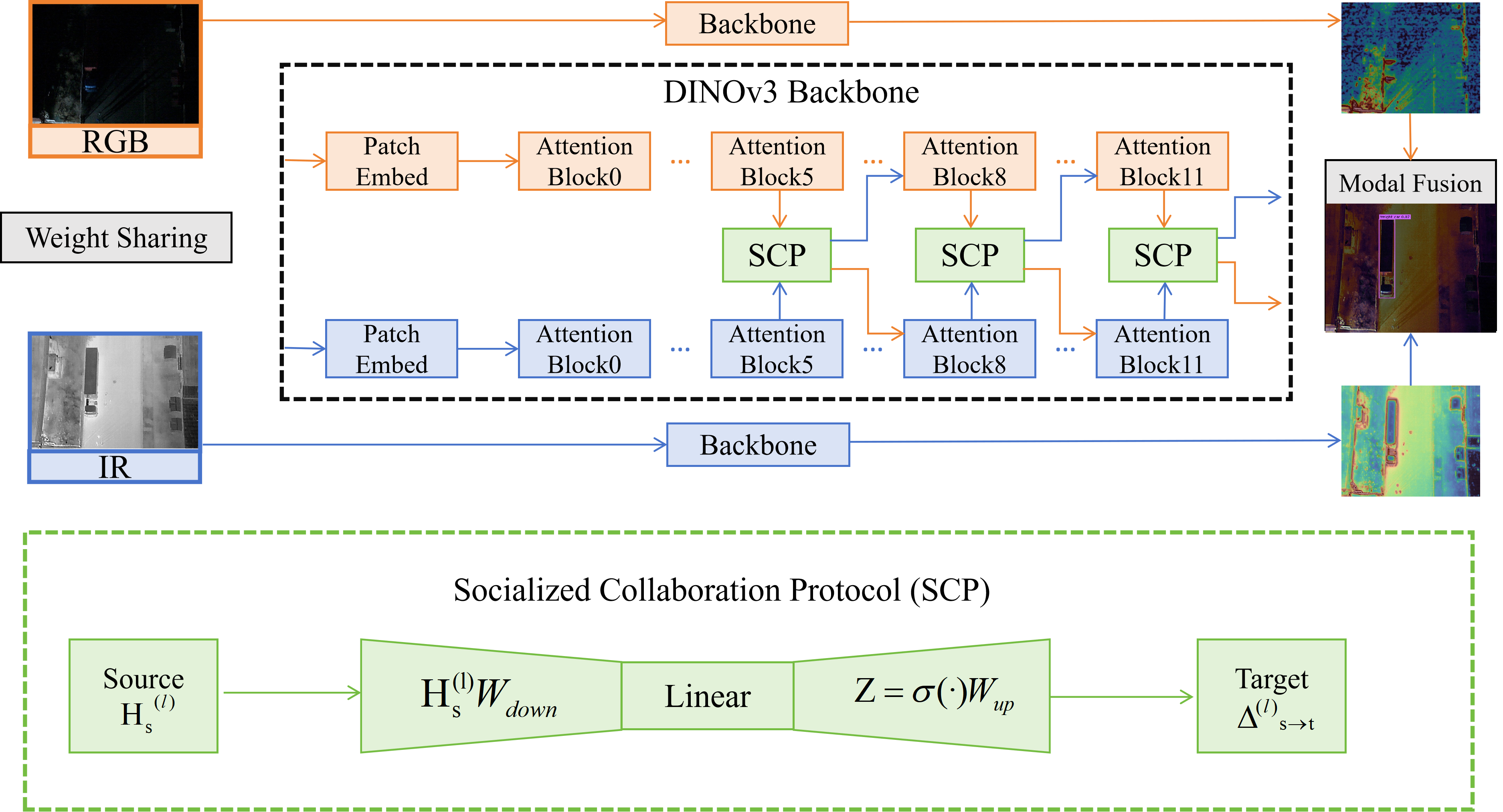}
    \caption{Overview of A2DINOv3. RGB and infrared images are encoded by two modality-specific streams with a parameter-shared DINOv3 backbone. Socialized Collaboration Protocols are inserted at selected layers to regulate bidirectional cross-modal communication. The resulting multi-level evidence is combined through parameter-free mean aggregation and forwarded to the detection head for joint prediction.}
    \label{fig:method}
\end{figure*}

Given a spatially paired RGB--infrared image pair
$\left(\bm{I}_{\mathrm{rgb}},\bm{I}_{\mathrm{ir}}\right)$,
the detector predicts a set of object instances
\begin{equation}
    \hat{\mathcal{Y}}
    =
    \left\{
    \left(
    \hat{\bm{b}}_{j},
    \hat{\bm{p}}_{j}
    \right)
    \right\}_{j=1}^{M},
\end{equation}
where $\hat{\bm{b}}_{j}$ and $\hat{\bm{p}}_{j}$ denote the bounding box and category distribution of the $j$-th prediction, respectively. The single-channel infrared image is replicated to match the backbone input format.

Multi-modal detection often assumes stronger interaction improves performance. Yet this fails when modalities are asymmetric. Dense fusion transmits both useful signals and modality-specific noise through the same path, allowing an unreliable modality to corrupt a reliable one.

Inspired by positive interdependence~\cite{co_a,co_b}, we argue that effective collaboration should preserve the competence of each expert while regulating their dependency. As illustrated in Fig.~\ref{fig:method}, A2DINOv3 implements this idea through three components: (1) a parameter-shared dual-stream backbone that maintains separate modality representations within a common semantic space; (2) SCP that constrains cross-modal exchange through low-dimensional residual pathways; and (3) zero initialization that keeps these pathways closed at the beginning of training and lets them emerge under task supervision. The framework therefore separates three roles often entangled in conventional fusion: representation, communication, and aggregation.

\subsection{Parameter-Shared Dual Experts}
\label{sec:dual_experts}

The RGB and infrared inputs are processed by two streams that share all DINOv3 backbone parameters but maintain separate activations. Parameter sharing transfers the semantic prior of the foundation model to both modalities and avoids duplicating the backbone. Separate feature propagation, in contrast, prevents heterogeneous observations from being mixed before their reliability can be assessed through task learning.

Let $\bm{X}_{m}^{(0)}$ denote the patch embeddings of modality
$m\in\{\mathrm{rgb},\mathrm{ir}\}$. At the $l$-th Transformer layer, the two streams are independently updated by the same backbone block:
\begin{equation}
    \bm{H}_{m}^{(l)}
    =
    \mathcal{B}_{l}
    \left(
    \bm{X}_{m}^{(l)};
    \bm{\theta}_{l}
    \right),
    \qquad
    m\in\{\mathrm{rgb},\mathrm{ir}\}.
    \label{eq:shared_backbone}
\end{equation}
where $\mathcal{B}_{l}$ and $\bm{\theta}_{l}$ denote the $l$-th Transformer block and its shared parameters.

\subsection{Socialized Collaboration Protocol}
\label{sec:scp}

The role of SCP is not to fully align the two modalities, but to control how much one expert can modify the representation of the other. At each interaction layer, two directional communication pathways are constructed for RGB-to-IR and IR-to-RGB transfer.

\paragraph{Bounded Cross-Modal Communication.}

For a source modality $s$ and a target modality $t$, the communication residual at layer $l$ is defined as
\begin{equation}
    \bm{\Delta}_{s\rightarrow t}^{(l)}
    =
    \sigma
    \left(
    \bm{H}_{s}^{(l)}
    \bm{W}_{\mathrm{down}}^{(l)}
    \right)
    \bm{W}_{\mathrm{up}}^{(l)},
    \label{eq:scp}
\end{equation}
where
$\bm{W}_{\mathrm{down}}^{(l)}
\in\mathbb{R}^{d\times r}$
and
$\bm{W}_{\mathrm{up}}^{(l)}
\in\mathbb{R}^{r\times d}$
are the down- and up-projection matrices, respectively. Here, $\sigma(\cdot)$ denotes GELU, and $r\ll d$ is the communication dimension.

The two experts are updated in parallel:
\begin{align}
    \bm{X}_{\mathrm{rgb}}^{(l+1)}
    &=
    \bm{H}_{\mathrm{rgb}}^{(l)}
    +
    \bm{\Delta}_{\mathrm{ir}\rightarrow\mathrm{rgb}}^{(l)},
    \\
    \bm{X}_{\mathrm{ir}}^{(l+1)}
    &=
    \bm{H}_{\mathrm{ir}}^{(l)}
    +
    \bm{\Delta}_{\mathrm{rgb}\rightarrow\mathrm{ir}}^{(l)}.
    \label{eq:bidirectional_scp}
\end{align}
Both residuals are computed from the pre-interaction features
$\bm{H}_{\mathrm{rgb}}^{(l)}$
and
$\bm{H}_{\mathrm{ir}}^{(l)}$.
The update is therefore symmetric and does not impose an artificial communication order between modalities.

For a non-interaction layer, the output is directly passed to the next layer:
\begin{equation}
    \bm{X}_{m}^{(l+1)}
    =
    \bm{H}_{m}^{(l)},
    \qquad
    l\notin\mathcal{I}.
\end{equation}
Cross-modal exchange is introduced only at a predefined set of layers,
\begin{equation}
    \mathcal{I}
    =
    \{5,8,11\},
\end{equation}
through explicit SCP pathways. This sparse placement preserves the backbone as the primary representation learner and confines modality interaction to a small number of controllable interfaces.

Unlike dense fusion, SCP restricts each cross-modal residual to a low-dimensional subspace:
\begin{equation}
    \operatorname{rank}
    \left(
    \bm{\Delta}_{s\rightarrow t}^{(l)}
    \right)
    \leq r.
\end{equation}
This constraint limits the degrees of freedom with which one modality can perturb the other. Since the available communication capacity is much smaller than the original feature dimension, the detection objective must allocate it to directions that consistently improve prediction. SCP therefore provides an inductive bias toward compact, task-relevant exchange rather than attempting to transmit every variation in the source modality.

Importantly, SCP does not assume that RGB or infrared is always more reliable. Both experts are subject to the same communication constraint, and the useful transfer directions are learned from data. The protocol thus accommodates sample-dependent modality quality without introducing manually defined reliability rules.

\begin{table*}[t]
\centering
\renewcommand{\arraystretch}{1.15}
\resizebox{\textwidth}{!}{%
\begin{tabular}{lccccccc}
\toprule
\multirow{2}{*}{\textbf{Method}} & \multirow{2}{*}{\textbf{Modality}} & \multicolumn{3}{c}{\textbf{Validation (\%)}} & \multicolumn{3}{c}{\textbf{Test (\%)}} \\
\cmidrule(lr){3-5} \cmidrule(lr){6-8}
& & \textbf{mAP} & \textbf{mAP$_{50}$} & \textbf{mAP$_{75}$} & \textbf{mAP} & \textbf{mAP$_{50}$} & \textbf{mAP$_{75}$} \\
\midrule
Faster R-CNN~\cite{Faster-RNN}   
& \multirow[c]{4}{*}{RGB}
& 35.97 & 63.35 & 36.59 & 8.50  & 25.04 & 3.46  \\

DDQ-DETR~\cite{DDQ}
&
& 40.55 & 66.77 & 44.05 & 7.46  & 22.24 & 2.86  \\
RF-DETR Large~\cite{RF-DETR}     &                         & 43.92 & 71.65 & 47.57 & \textbf{10.15} & \textbf{29.45} & 3.79  \\
YOLO26-X~\cite{YOLO26}           &                         & \textbf{49.85} & \textbf{77.25} & \textbf{56.24} & 9.76  & 27.82 & \textbf{4.21}  \\
\midrule
Faster R-CNN~\cite{Faster-RNN}   &     \multirow{4}{*}{IR}     & 49.40 & 71.95 & 59.45 & 28.47 & 44.46 & 32.25 \\
DDQ-DETR~\cite{DDQ}             &      & 52.63 & 73.42 & 63.71 & 25.92 & 36.85 & 30.53 \\
RF-DETR Large~\cite{RF-DETR}     &                         & 54.40 & 75.26 & 65.59 & \textbf{37.12} & \textbf{53.40} & \textbf{44.34} \\
YOLO26-X~\cite{YOLO26}           &                         & \textbf{62.63} & \textbf{83.74} & \textbf{74.78} & 36.07 & 49.97 & 42.63 \\
\midrule
CSAA~\cite{CSAA}                 & \multirow{5}{*}{RGB+IR} & 34.50 & 50.72 & 41.87 & 16.72 & 27.99 & 18.16 \\
ICAFusion~\cite{ICAFusion}       &                         & 60.43 & 81.05 & 72.82 & 35.62 & 50.70 & 42.31 \\
M\textsuperscript{2}D-LIF~\cite{M2D} &                       & 60.52 & 79.95 & 72.44 & 34.76 & 48.90 & 40.83 \\
AFF-Net~\cite{AFFNET}            &                         & 60.32 & 81.22 & 72.78 & 37.53 & 56.05 & 43.38 \\
\textbf{A2DINOv3 (Ours)}         &                         & \textbf{64.00} & \textbf{84.28} & \textbf{76.21} & \textbf{43.05} & \textbf{61.08} & \textbf{50.56} \\
\bottomrule
\end{tabular}%
}
\caption{Object detection results on the GAIIC dataset. Comparison of our proposed A2DINOv3 with baselines on Validation and Test sets. All metrics are reported in percentage (\%). The best results within each modality group are highlighted in bold.}
\label{tab:main_gaiic}
\vspace{-15pt}
\end{table*}

\paragraph{Progressive Collaboration via Zero Initialization.}

Although the bottleneck restricts communication capacity, randomly initialized residuals may still perturb the pre-trained feature space at the beginning of optimization. We therefore initialize every up-projection matrix as
\begin{equation}
    \bm{W}_{\mathrm{up}}^{(l)}
    =
    \bm{0}.
\end{equation}
At initialization, the communication residual is exactly zero:
\begin{equation}
    \bm{\Delta}_{s\rightarrow t}^{(l)}
    =
    \bm{0}.
\end{equation}
Consequently, the gradient contributed by this communication path to the source representation and the down-projection matrix also vanishes:
\begin{equation}
    \left.
    \frac{\partial\mathcal{L}}
    {\partial\bm{H}_{s}^{(l)}}
    \right|_{\mathrm{SCP}}
    =
    \bm{0},
    \qquad
    \frac{\partial\mathcal{L}}
    {\partial\bm{W}_{\mathrm{down}}^{(l)}}
    =
    \bm{0}.
    \label{eq:zero_grad}
\end{equation}
The first term in Eq.~\eqref{eq:zero_grad} refers only to the gradient transmitted through SCP; the source stream still receives gradients through its ordinary detection pathway.

Critically, the communication pathway remains trainable because the gradient w.r.t. the up-projection matrix,
\begin{equation}
    \frac{\partial\mathcal{L}}
    {\partial\bm{W}_{\mathrm{up}}^{(l)}}
    =
    \sigma
    \left(
    \bm{H}_{s}^{(l)}
    \bm{W}_{\mathrm{down}}^{(l)}
    \right)^{\top}
    \frac{\partial\mathcal{L}}
    {\partial\bm{\Delta}_{s\rightarrow t}^{(l)}},
\end{equation}
is generally non-zero. Thus, $\bm{W}_{\mathrm{up}}$ receives updates from the first optimization step, after which the non-zero residual gradually enables gradients to flow back to both the down-projection matrix and the source representation through the SCP path.

This zero-initialization induces a progressive optimization process. At the early stage of training, the shared backbone operates as two independent experts, preserving its pre-trained representations. Cross-modal dependency emerges only when the learned communication residuals contribute positively to the detection loss. Collaboration therefore grows from an identity-preserving state, rather than being imposed through random feature perturbations.

\subsection{Collaborative Evidence Aggregation}
\label{sec:aggregation}
After bounded interaction, the two streams retain their modality-specific representations while incorporating compact evidence from the other modality. The selected backbone outputs at interaction layers $l\in\mathcal{I}$ are converted into multi-level feature maps $\bm{F}_{\mathrm{rgb}}^{(l)}$
and
$\bm{F}_{\mathrm{ir}}^{(l)}$,
which are aggregated via parameter-free mean fusion:
\begin{equation}
    \bm{F}_{\mathrm{fused}}^{(l)}
    =
    \frac{1}{2}
    \left(
    \bm{F}_{\mathrm{rgb}}^{(l)}
    +
    \bm{F}_{\mathrm{ir}}^{(l)}
    \right),
    \qquad
    l\in\mathcal{I}.
    \label{eq:fusion}
\end{equation}

This simple aggregation is intentional: SCP determines what information is exchanged, while fusion merely collects the resulting evidence. Introducing a high-capacity fusion module would blur this separation and risk reintroducing unrestricted cross-modal mixing. Mean fusion avoids additional modality-dependent parameters and preserves balanced contributions from both streams.

The fused feature pyramid $\{\bm{F}_{\text{fused}}^{(l)}\}_{l\in\mathcal{I}} $comprising the aggregated representations from all interaction layers—is passed to a hybrid encoder and a DETR decoder. Following standard set-based detection, Hungarian matching assigns predictions to ground-truth objects. The training objective is
\begin{equation}
    \mathcal{L}
    =
    \mathcal{L}_{\mathrm{cls}}
    +
    \lambda_{\mathrm{L1}}
    \mathcal{L}_{\mathrm{L1}}
    +
    \lambda_{\mathrm{giou}}
    \mathcal{L}_{\mathrm{giou}},
    \label{eq:detection_loss}
\end{equation}
where $\mathcal{L}_{\mathrm{cls}}$ is the classification loss, and $\mathcal{L}_{\mathrm{L1}}$ and $\mathcal{L}_{\mathrm{giou}}$ are the $L_{1}$ and generalized IoU losses for bounding-box regression.

\section{Experiments}
\label{sec:experiments}

We evaluate A2DINOv3 on four widely used RGB--infrared object detection benchmarks, including GAIIC2024~\cite{GAIIC2024}, LLVIP~\cite{llvip}, FLIR~\cite{FLIR}, and M3FD~\cite{m3fd}. Our experiments are designed to answer three questions: 
(1) whether A2DINOv3 improves multi-modal detection across diverse scenarios;
(2) whether bounded communication alleviates fusion degradation caused by dense cross-modal interaction; and
(3) whether zero initialization enables stable adaptation of the pre-trained foundation model.
All experiments are implemented with PyTorch and conducted on eight NVIDIA RTX 4090 GPUs.

\subsection{Experimental Setup}
\label{sec:exp_implementation}

\paragraph{Datasets and Metrics.}
GAIIC2024 contains challenging aerial scenes with substantial illumination variation and complex backgrounds. FLIR focuses on autonomous-driving scenarios, while LLVIP primarily covers nighttime surveillance scenes. M3FD further includes diverse urban environments and adverse imaging conditions. Following the standard protocols of each benchmark, we report mean Average Precision (mAP), including mAP$_{50}$ and mAP$_{75}$ when available.

\paragraph{Compared Methods.}
We compare A2DINOv3 with representative single-modal and multi-modal detectors. The single-modal baselines include Faster R-CNN~\cite{Faster-RNN}, DDQ-DETR~\cite{DDQ}, RF-DETR~\cite{RF-DETR}, and YOLO26-X~\cite{YOLO26}. The multi-modal competitors include CSAA~\cite{CSAA}, ICAFusion~\cite{ICAFusion}, M\textsuperscript{2}D-LIF~\cite{M2D}, and AFF-Net~\cite{AFFNET}. Unless otherwise specified, predicted boxes with confidence scores below 0.25 are removed during evaluation. Additional training configurations and dataset-specific settings are provided in the Appendix.

\begin{figure}[t]
    \centering
    \includegraphics[width=\columnwidth]{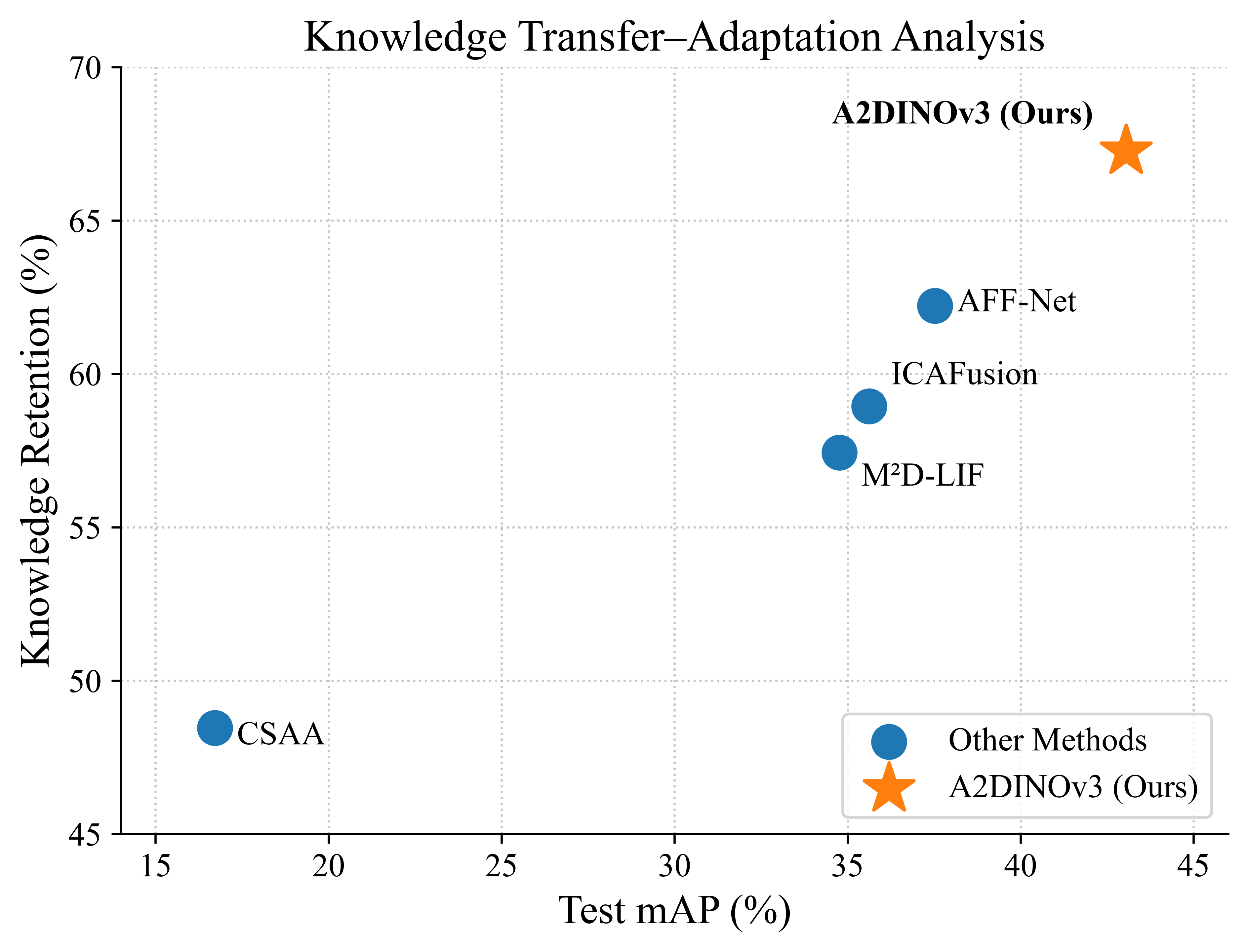}
    \caption{Knowledge retention analysis on the GAIIC2024 test set. A2DINOv3 achieves a favorable balance between detection accuracy and knowledge preservation.}
    \label{fig:retention}
    \vspace{-10pt}
\end{figure}

\subsection{Main Results on GAIIC2024}
\label{sec:main_results}
Table~\ref{tab:main_gaiic} reports the results on GAIIC2024. A2DINOv3 achieves the best multimodal performance, reaching 64.00\% and 43.05\% mAP on the validation and test sets, respectively. On the test set, it surpasses AFF-Net by 5.52\%, 5.03\%, and 7.18\% in mAP, mAP$_{50}$, and mAP$_{75}$.

The results reveal a clear modality asymmetry: RGB-only models degrade significantly under challenging conditions, whereas IR-only models remain substantially stronger. Moreover, several RGB--IR methods underperform the strongest IR-only baseline, indicating that naive multimodal fusion may introduce negative transfer when cross-modal interactions are not properly controlled.

In contrast, A2DINOv3 improves over the strongest IR-only baseline by 5.93\% mAP on the test set, demonstrating that degraded RGB observations still provide valuable complementary information. By constraining cross-modal communication, SCP selectively incorporates informative RGB cues while preserving reliable infrared representations.

The performance gap between validation and test sets further reflects the difficulty of UAV-based cross-domain detection. Compared with the validation set, the test set contains more distant views, smaller targets, and larger distribution variations, leading to considerable performance degradation for existing methods. As shown in Fig.~\ref{fig:retention}, A2DINOv3 achieves the highest test mAP while maintaining the best knowledge retention among multimodal methods, preserving 67.27\% of its validation performance. This result verifies that SCP facilitates effective knowledge transfer across modalities and maintains stable representations under challenging UAV scenarios.

\subsection{Cross-Domain Generalization}
\label{sec:generalization}

\begin{table}[htbp]
\centering
\small
\renewcommand{\arraystretch}{1.15}
\setlength{\tabcolsep}{4pt}

\begin{tabular}{l c c c c}
\toprule
\textbf{Method} & \textbf{Modality} & 
\textbf{M3FD} & \textbf{FLIR} & \textbf{LLVIP}\\
\midrule

Faster R-CNN 
& \multirow[c]{4}{*}{RGB}
& 47.86 & 28.90 & 45.10 \\

DDQ-DETR
& & 34.62 & 30.90 & 46.70 \\

RF-DETR Large
& & \textbf{54.44} & \textbf{38.07} & \textbf{57.24} \\

YOLO26-X
& & 29.71 & 30.22 & 48.37 \\

\midrule

Faster R-CNN
& \multirow[c]{4}{*}{IR}
& 41.98 & 37.60 & 54.50 \\

DDQ-DETR
& & 30.88 & 37.10 & 58.60 \\

RF-DETR Large
& & \textbf{52.05} & \textbf{46.57} & \textbf{70.35} \\

YOLO26-X
& & 17.06 & 39.98 & 66.00 \\

\midrule

CSAA
& \multirow[c]{5}{*}{RGB+IR}
& 46.32 & 41.30 & 59.20 \\

ICAFusion
& & 59.09 & 41.40 & 64.30 \\

M\textsuperscript{2}D-LIF
& & 55.79 & 46.10 & 70.80 \\

AFF-Net
& & 59.22 & 41.84 & 65.15 \\

\textbf{A2DINOv3 (Ours)}
& &
\textbf{61.78} &
\textbf{46.80} &
\textbf{71.56} \\

\bottomrule
\end{tabular}

\caption{Comparison with state-of-the-art methods on M3FD, FLIR, and LLVIP datasets. Results are reported in mAP (\%).}
\label{tab:generalization}
\end{table}

We evaluate A2DINOv3 on M3FD, FLIR, and LLVIP, which cover urban perception, autonomous driving, and nighttime surveillance. As reported in Tables~\ref{tab:generalization}, A2DINOv3 achieves 61.78\%, 46.80\%, and 71.56\% mAP, respectively, demonstrating consistent generalization across diverse domains. The FLIR results further reveal the risk of unregulated fusion: AFF-Net and ICAFusion achieve 41.84\% and 41.40\% mAP, both below the 46.57\% of the IR-only RF-DETR. In contrast, A2DINOv3 reaches 46.80\% by preserving reliable infrared representations while selectively incorporating useful RGB cues. These results suggest that its robustness stems from regulating both the capacity and optimization of cross-modal interaction, rather than learning dataset-specific fusion patterns.

\subsection{Ablation Study}
\label{sec:ablation}

We conduct ablation studies on GAIIC2024 to examine the effects of the communication bottleneck and optimization strategy. As shown in Table~\ref{tab:ablation}, the infrared-only baseline achieves 36.47\% mAP.

\begin{table}[htbp]
\centering
\setlength{\tabcolsep}{4pt}
\renewcommand{\arraystretch}{1.15}
\begin{tabular}{llc}
\toprule
\textbf{Interaction} & \textbf{Optimization} & \textbf{mAP (\%)} \\
\midrule
None
& IR-only
& 36.47 \\
\midrule
\multirow{3}{*}{Cross-Attn}
& Layers 5/8/11
& 38.02 \\
& Frozen Backbone
& 38.10 \\
& Full Fine-tuning
& 39.30 \\
\midrule
\multirow{3}{*}{SCP}
& Frozen Backbone
& 40.27 \\
& Full Fine-tuning (Random Init.)
& 42.43 \\
& \textbf{Full Fine-tuning (Zero Init.)}
& \textbf{43.05} \\
\bottomrule
\end{tabular}
\caption{Ablation study on GAIIC2024. We compare different cross-modal interaction mechanisms and backbone optimization strategies.}
\label{tab:ablation}
\vspace{-15pt}
\end{table}
\paragraph{Effect of Bounded Communication.}
With the backbone frozen, replacing dense cross-attention with SCP improves mAP from 38.10\% to 40.27\%. Since both variants use the same frozen backbone, the 2.17-point gain mainly reflects the difference in cross-modal interaction. Dense cross-attention provides unrestricted access to the source features, whereas SCP constrains the exchanged residual to a low-dimensional subspace. The result supports the hypothesis that limiting communication capacity is beneficial when modality quality is asymmetric.

\begin{figure*}[!htbp]
\centering
\includegraphics[width=0.85\textwidth]{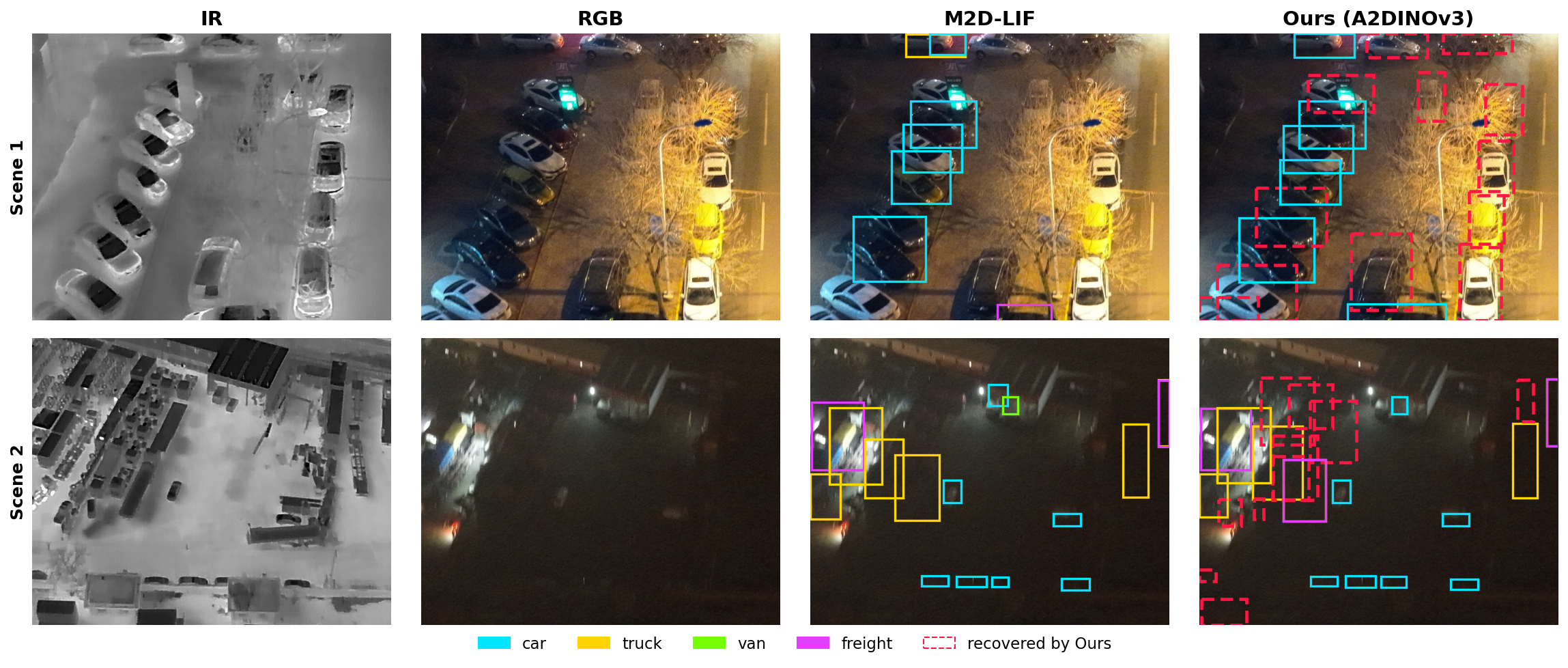}
\vspace{-10pt}
\caption{Qualitative comparison between M\textsuperscript{2}D-LIF and A2DINOv3 under extreme low-light and complex illumination conditions. Red dashed circles highlight representative missed or recovered targets.}
\label{fig:qualitative}
\vspace{-15pt}
\end{figure*}

\paragraph{Effect of Optimization Strategy and Initialization.}
Full fine-tuning raises cross-attention from 38.10\% to 39.30\% mAP, but still underperforms frozen SCP. In contrast, full fine-tuning with SCP significantly boosts performance. While SCP with random initialization achieves 42.43\% mAP, our zero-initialization strategy further elevates the accuracy to 43.05\%. This specific gain confirms that zero-initialization effectively suppresses initial cross-modal gradient shocks and gradually activates communication, enabling stable joint optimization of the backbone and SCP.

\subsection{Qualitative Results}
\label{sec:visualization}

Fig~\ref{fig:qualitative} compares A2DINOv3 with M\textsuperscript{2}D-LIF under severe low-light and uneven illumination. M\textsuperscript{2}D-LIF misses several shadowed or cluttered targets, revealing the vulnerability of dense interaction to degraded RGB features. In contrast, A2DINOv3 recovers more difficult targets, showing that bounded communication preserves complementary cues while suppressing cross-modal interference.

\begin{figure}[htbp]
\vspace{-10pt}
\centering
\includegraphics[width=\columnwidth]{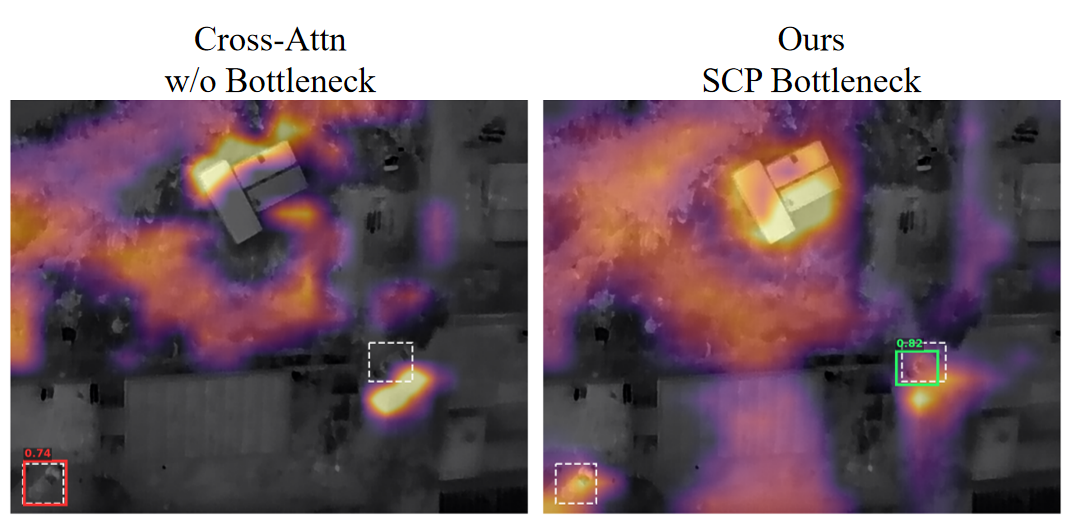}
\vspace{-15pt}
\caption{Feature responses of dense cross-attention and A2DINOv3 under degraded RGB inputs. Bounded communication produces more target-focused activations.}
\label{fig:mechanistic_analysis}
\vspace{-15pt}
\end{figure}

\subsection{Mechanistic Analysis}
\label{sec:mechanistic}

To further analyze why A2DINOv3 improves collaboration, we investigate the internal behavior from two perspectives: cross-modal feature interaction and optimization dynamics.

\paragraph{Feature Responses under Bounded Communication.}
As shown in Fig.~\ref{fig:mechanistic_analysis}, dense cross-attention produces dispersed activations and a false positive with 0.74 confidence. In contrast, A2DINOv3 concentrates its response on the true target and yields a correct prediction with 0.82 confidence, suggesting that low-dimensional communication reduces sensitivity to irrelevant RGB responses.

\begin{figure}[htbp]
\centering
\includegraphics[width=\columnwidth]{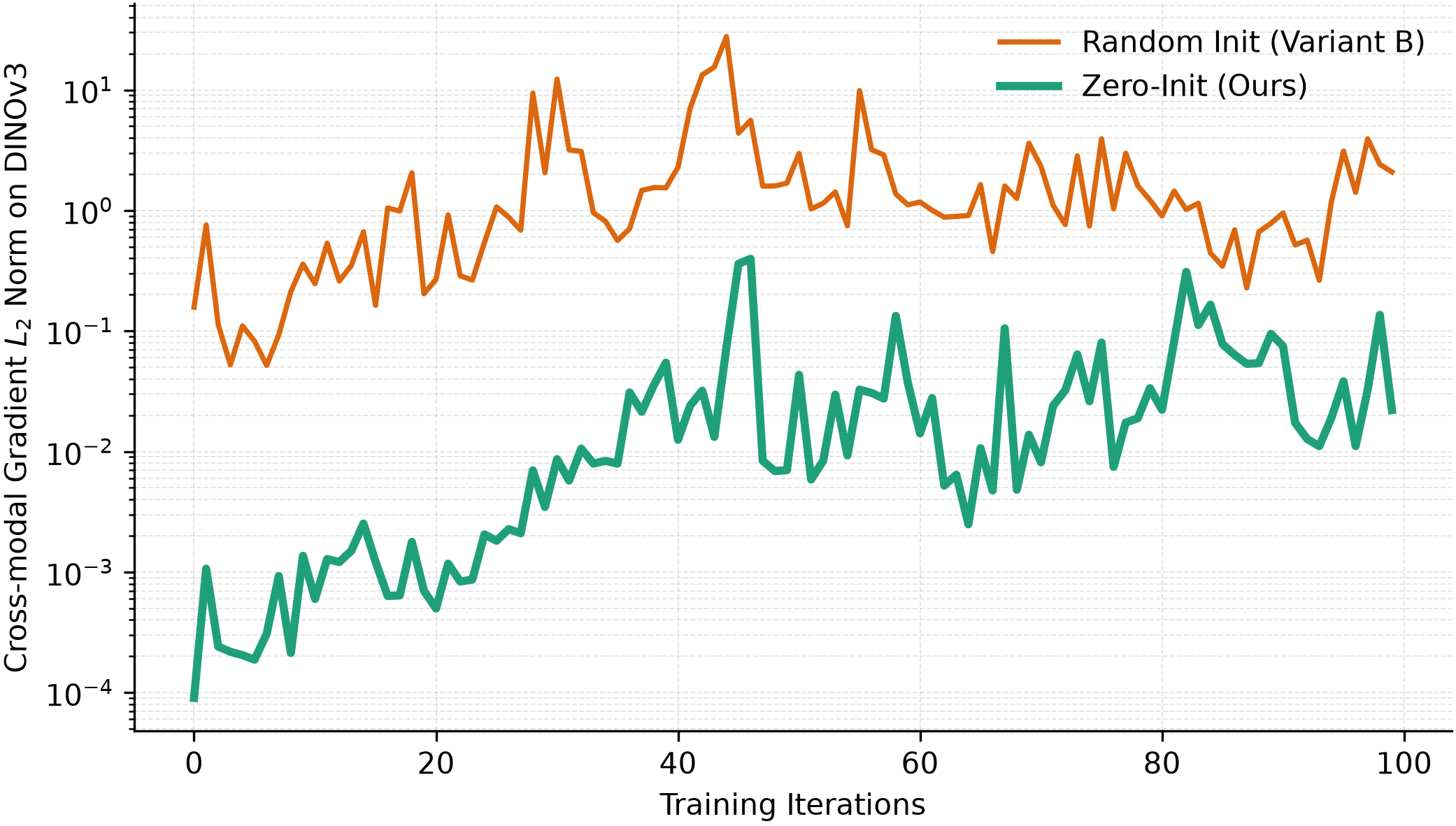}
\vspace{-15pt}
\caption{Cross-modal gradient dynamics during early training. Zero initialization suppresses the initial gradient and progressively activates communication.}
\label{fig:zero_init_gradient}
\vspace{-15pt}
\end{figure}

\paragraph{Optimization Dynamics under Zero Initialization.}
We track the $L_2$ norm of cross-modal gradients during the first 100 iterations. Fig.~\ref{fig:zero_init_gradient} shows that random initialization yields an initial magnitude near $10^{1}$, whereas zero initialization suppresses it to approximately $10^{-4}$. The gradient then increases gradually as the up-projection matrices depart from zero, confirming that communication is progressively activated rather than permanently disabled. This behavior is consistent with the collaboration mechanism in SCP section.

\section{Conclusion}
\label{sec:conclusion}

This work addresses two challenges in adapting vision foundation models to multimodal detection: fusion degradation from unreliable cross-modal signals and distortion of pre-trained representations caused by abrupt joint optimization. Inspired by Social Interdependence Theory, we attribute these issues to unrestricted communication and premature dependency between heterogeneous modalities. We therefore propose A2DINOv3, which models RGB and IR as complementary experts and regulates their interaction via SCP with zero-initialized pathways, enabling a gradual transition from independent learning to coordinated collaboration. Our results show that effective multimodal adaptation depends not on maximizing feature exchange, but on regulating when, what, and how much information is communicated.


\clearpage

\appendix

\section{Appendix }
The appendix contains comprehensive details on the implementations and experimental results referenced in the main paper, along with supplementary theoretical analysis and in-depth discussions. It is organized as follows:
\label{A}

\begin{itemize}
    \item In Implementation Details , we offer a thorough overview of the methods compared in the main paper, accompanied by a detailed description of the datasets utilized.
    \item In Method Details, we detail the formal mathematical background of Social Interdependence Theory and present the complete algorithmic pipeline to illustrate the execution of A2DINOv3.
    \item In Full Experimental Results, we present the full set of experimental results, accompanied by an in-depth analysis that thoroughly evaluates the model's performance.
\end{itemize}

\section{Implementation Details}

In this section, we offer a detailed description of the methods compared in the main paper, as well as the datasets used. 

We evaluate our models on three RGB-T benchmarks: GAIIC2024, FLIR, and LLVIP, under a unified COCO-style protocol with a confidence threshold of 0.25 and maxDets=[10,100,500]. Our A2DINOv3 framework utilizes dual DINOv3 ViT-S/16+ backbones (interaction indexes [5, 8, 11]) and a 6-layer DEIM decoder with 300 object queries. The model is optimized using AdamW ($\beta=(0.9,0.999)$) with a weight decay of $1.25 \times 10^{-4}$, mixed-precision training, EMA, and gradient clipping of 0.1.

For GAIIC2024, the model is trained for 58 epochs using a total batch size of 64. The base learning rate is set to $5 \times 10^{-4}$ (backbone $1 \times 10^{-5}$) with a flat-cosine scheduler. Images are resized to $640 \times 640$, and the training incorporates robust data augmentations including Mosaic, mixup, random photometric distortion, and CopyBlend. For FLIR, the model is fine-tuned for 36 epochs with a batch size of 8 and a base learning rate of $2 \times 10^{-4}$, utilizing the same augmentations. For LLVIP, the model is fine-tuned for 8 epochs with a batch size of 8 and a base learning rate of $1 \times 10^{-4}$ at a resolution of $768 \times 768$, with strong augmentations disabled. Furthermore, the YOLO26-X baselines are trained for 50 epochs with a batch size of 24 and an image size of 640.

\subsection{ Compared Methods}

In this subsection, we provide an overview of the methods compared in the main paper, outlining their key characteristics. The methods considered are as follows:

\begin{itemize}
    \item \textbf{Faster R-CNN ~\cite{Faster-RNN}:} Introduces a Region Proposal Network (RPN) that shares full-image convolutional features with the detection network, unifying both into a single architecture to enable efficient, nearly cost-free region proposals.
    
    \item \textbf{DDQ-DETR ~\cite{DDQ}:}Introduces a Dense Distinct Query (DDQ) mechanism that employs a distinct queries selection pre-processing step to filter out similar proposals, mitigating optimization difficulties and accelerating convergence in end-to-end detectors 
    
    \item \textbf{YOLO26-X~\cite{YOLO26}:}Proposes a unified real-time vision model family that utilizes a dual-head design for native NMS-free end-to-end inference and removes the Distribution Focal Loss (DFL) to yield a lighter, unconstrained regression head.
    
    \item \textbf{RF-DETR~\cite{RF-DETR}:} Introduces a light-weight specialist detection transformer that leverages weight-sharing neural architecture search (NAS) to discover optimal accuracy-latency Pareto curves for target domains without the need for retraining.
    
    \item \textbf{CSAA~\cite{CSAA}:} Proposes a lightweight multimodal fusion module that employs channel switching and parameter-free spatial attention to efficiently integrate cross-modal features, significantly improving detection accuracy without introducing excessive computational overhead.

    \item \textbf{AFF-Net~\cite{AFFNET}:} Proposes an adaptive fine-grained fusion network that leverages a local feature consistency-based module to dynamically assign fusion weights, and introduces a mutual information-guided contrastive loss to preserve modality-specific features, effectively handling complex illumination and dense occlusions in UAV scenarios.

    \item \textbf{ICAfusion~\cite{ICAFusion}:} Proposes a dual cross-attention transformer framework to model global feature interactions across modalities, incorporating an iterative learning strategy that shares parameters across blocks to continuously refine complementary features without increasing model complexity.

    \item \textbf{M\textsuperscript{2}D-LIF~\cite{M2D}:}Rethinks multi-modal object detection from a mono-modality learning perspective by introducing a Mono-Modality Distillation (M$^2$D) method to ensure sufficient feature learning during joint training, paired with a Local Illumination-aware Fusion (LIF) module that dynamically weights features based on illumination conditions, thereby effectively mitigating the Fusion Degradation phenomenon.
\end{itemize}

\subsection{ Datasets}

Our experiments are conducted on four standard RGB-T object detection benchmarks, each chosen to represent a different level of environmental complexity and sensor heterogeneity.

\begin{itemize}
    \item \textbf{GAIIC2024 ~\cite{GAIIC2024}:} This dataset is a specialized RGB-TIR aerial object detection benchmark captured from unmanned aerial vehicles, featuring diverse scenes such as urban roads and residential areas with challenging lighting conditions and frequent image misalignments. The dataset encompasses five vehicle categories, and the challenge task aimed to leverage complementary RGB and TIR information to improve detection robustness.

    \item \textbf{FLIR ~\cite{FLIR}:} This dataset provides 5,142 aligned visible-infrared paired images, including day and night scenes with three object categories: person, car, and bicycle. The dataset is widely utilized for multispectral object detection and fusion research, though it originally contained misaligned pairs that necessitate careful filtering for training. The images are particularly valuable for evaluating detection robustness under varying illumination conditions.

    \item \textbf{LLVIP ~\cite{llvip}:} This dataset is a large-scale visible-infrared paired benchmark specifically designed for low-light vision tasks, containing 33,672 images (16,836 pairs) predominantly captured in extremely dark scenes. All image pairs are strictly aligned in both time and space, with annotated pedestrian labels, making it highly effective for evaluating multispectral pedestrian detection and image fusion algorithms under challenging lighting conditions where detail is otherwise lost.

    \item \textbf{M3FD~\cite{m3fd}:} This dataset is a comprehensive multi-scene multi-modality benchmark providing 4,200 aligned visible and infrared image pairs captured under various environmental conditions, including daytime, overcast, and night scenarios. It encompasses six object categories: people, car, bus, motorcycle, truck, and lamp. The rich diversity of scenes and illumination variations makes it particularly valuable for evaluating the generalization ability and robustness of multi-modal fusion algorithms across complex real-world environments.
    
\end{itemize}

\section{ Method Details}

\subsection{Theoretical Background}
\label{sec:interdependence_theory}

Social interdependence theory, originally proposed by Morton Deutsch in 1949, explores how the structured association of goals among individuals in a group determines their interaction patterns and ultimate outcomes. Within the frameworks of mathematical logic and probability theory, this theory can be strictly formalized through the conditional probabilities of individual goal attainment.

Assume there are two interacting entities, $\mathcal{A}$ and $\mathcal{B}$, within a collaborative system. Let $G_A$ and $G_B$ represent the events that entity $\mathcal{A}$ and entity $\mathcal{B}$ successfully achieve their respective goals. Meanwhile, let $P(G_A)$ and $P(G_B)$ denote their marginal probabilities of goal attainment (i.e., their independent success rates without considering each other's state).

Based on the intrinsic nature of the goal structure, Social Interdependence Theory categorizes the dependency between entities into the following three fundamental forms.

\textbf{Positive interdependence:} This state of collaboration describes a cooperative goal structure. In this state, the success of one entity facilitates the success of the other, exhibiting a positive logical alignment in goal attainment. Probabilistically, this means that given the condition that entity $\mathcal{B}$ achieves its goal, the conditional probability of entity $\mathcal{A}$ achieving its goal is significantly greater than its marginal probability. This is formalized as:
{\small
\begin{equation}
P(G_A | G_B) > P(G_A) \quad \mathrm{and} \quad P(G_B | G_A) > P(G_B)
\end{equation}
}
In this structure, individuals tend to engage in promotive interaction, maximizing joint benefits through resource sharing and information complementarity.

\textbf{Negative interdependence:} In this state, there exists a goal structure among collaborators that is competitive or mutually interfering. Here, the success of one entity is predicated on the failure of the other, or the actions of one entity substantially hinder the progress of the other. Formally, the conditional probability of goal attainment is less than the independent marginal probability:
{\small
\begin{equation}
P(G_A | G_B) < P(G_A) \quad \mathrm{and} \quad P(G_B | G_A) < P(G_B)
\end{equation}
}
In this structure, individuals often engage in contrient interaction, leading to internal friction, information blocking, and mutual interference within the system.

\textbf{No interdependence:} It represents a completely independent goal structure. In this scenario, the actions and outcomes of any one entity neither interfere with nor facilitate the others. Mathematically, this is equivalent to the two goal events being statistically independent:
{\small
\begin{equation}
P(G_A | G_B) = P(G_A) \quad \mathrm{and} \quad P(G_B | G_A) = P(G_B)
\end{equation}
}
Individuals in a state of no interdependence execute tasks completely independently. The system manifests as a simple superimposition of isolated modules, lacking synergistic gains.

\noindent Through these formal definitions, Social Interdependence Theory provides a rigorous mathematical perspective for analyzing collaborative behaviors in complex systems. It demonstrates that constructing an efficient collaborative system relies on establishing structured boundary conditions for positive interdependence, thereby physically or logically preventing the negative interdependence degradation caused by unconstrained interactions.

\subsection{Algorithm}
To provide a more comprehensive understanding of the operational flow of A2DINOv3, Algorithm \ref{alg:a2dinov3} details the complete training and inference pipeline. During the initialization phase, the RGB and infrared streams are set to share the pre-trained weights of the DINOv3 backbone. Crucially, the up-projection matrices $\bm{W}_{up}^{(l)}$ in all Socialized Collaboration Protocols (SCPs) are strictly zero-initialized to buffer the initial cross-modal gradient shocks. 

In the training forward pass, both modalities undergo independent feature extraction at non-interaction layers. At the predefined interaction layers $l \in \mathcal{I}$, the complementary modality's features are compressed and projected through the SCP bottleneck to generate bounded cross-modal residuals $\bm{\Delta}_{\bar{m}\rightarrow m}^{(l)}$, which are then injected into the target modality. This mechanism ensures selective information exchange while rigorously preserving modality-specific competence. Subsequently, the refined multi-level features from both streams are aggregated via parameter-free mean fusion. The fused representations are then fed into the DETR decoder to predict bounding boxes and classification scores, followed by joint optimization via back-propagation. During the inference phase, the network executes a single forward pass, yielding the final detection results by filtering the predictions with a confidence threshold of 0.25.

\begin{algorithm}[H]
\caption{Training and Inference of A2DINOv3}
\label{alg:a2dinov3}
\begin{algorithmic}[1]

\REQUIRE Dataset $\mathcal{D}$ with annotations $\bm{G}$; DINOv3 backbone $\bm{\theta}_{\text{DINO}}$; epochs $T$; interaction layers $\mathcal{I}=\{5,8,11\}$.
\STATE \textbf{Initialize:} Share $\bm{\theta}_{\text{DINO}}$ across RGB/IR streams; set $\bm{W}_{up}^{(l)}=\bm{0}$ for all SCPs.

\FOR{$t=1$ \TO $T$, and each batch $(\bm{I}_{rgb},\bm{I}_{ir},\bm{G})\in\mathcal{D}$}
    \STATE Extract initial patch embeddings $\bm{X}_{m}^{(0)}$ for modality $m \in \{rgb, ir\}$.
    
    \FOR{layer $l=0$ \TO $L-1$}
        \STATE Forward block: $\bm{H}_{m}^{(l)} = \mathcal{B}_l(\bm{X}_{m}^{(l)}; \bm{\theta}_l)$.
        \IF{$l\in\mathcal{I}$}
            \STATE SCP: $\bm{\Delta}_{\bar{m}\rightarrow m}^{(l)} = \sigma(\bm{H}_{\bar{m}}^{(l)}\bm{W}_{down}^{(l)})\bm{W}_{up}^{(l)}$ \quad \COMMENT{$\bar{m}$ is the complementary modality}
            \STATE Update features: $\bm{X}_{m}^{(l+1)} = \bm{H}_{m}^{(l)} + \bm{\Delta}_{\bar{m}\rightarrow m}^{(l)}$
        \ELSE
            \STATE Identity pass: $\bm{X}_{m}^{(l+1)} = \bm{H}_{m}^{(l)}$
        \ENDIF
    \ENDFOR
    
    \STATE \textbf{Aggregate:} Mean fusion $\bm{F}_{fused}^{(l)} = \frac{1}{2}(\bm{X}_{rgb}^{(l+1)} + \bm{X}_{ir}^{(l+1)})$ for $l \in \mathcal{I}$.
    \STATE \textbf{Predict:} Bounding boxes and classes $\{(\hat{\bm{b}}, \hat{\bm{p}})\} \leftarrow \text{DETR}(\{\bm{F}_{fused}^{(l)}\})$.
    \STATE Compute joint loss $\mathcal{L}$ with $\bm{G}$ and update parameters via back-propagation.
\ENDFOR

\STATE \textbf{Inference:} Forward test pair $(\bm{I}_{rgb}, \bm{I}_{ir})$, return $\{(\hat{\bm{b}}, \hat{\bm{p}})\}$ filtered by score $\ge 0.25$.

\end{algorithmic}
\end{algorithm}

\section{ Full Experimental Results}

\begin{table}[htbp]
\centering
\small
\setlength{\tabcolsep}{3.5pt} 
\begin{tabular}{llccc}
\toprule
\textbf{Method} & \textbf{Modality} & \textbf{mAP} & \textbf{mAP$_{50}$} & \textbf{mAP$_{75}$ } \\
\midrule
Faster R-CNN              & \multirow{4}{*}{RGB}    & 47.86 & 77.72 & 49.37 \\
DDQ-DETR                  &                         & 34.62 & 63.08 & 33.58 \\
RF-DETR Large             &                         & \textbf{54.44} & \textbf{82.87} & \textbf{56.95} \\
YOLO26-X                  &                         & 29.71 & 51.40 & 29.82 \\
\midrule
Faster R-CNN              & \multirow{4}{*}{IR}     & 41.98 & 67.36 & 43.83 \\
DDQ-DETR                  &                         & 30.88 & 55.94 & 31.89 \\
RF-DETR Large             &                         & \textbf{52.05} & \textbf{80.64} & \textbf{53.12} \\
YOLO26-X                  &                         & 17.06 & 30.87 & 17.37 \\
\midrule
CSAA                      & \multirow{5}{*}{RGB+IR} & 46.32 & 76.46 & 48.62 \\
ICAFusion                 &                         & 59.09 & 88.43 & 63.72 \\
M\textsuperscript{2}D-LIF &                         & 55.79 & 81.11 & 60.18 \\
AFF-Net                   &                         & 59.22 & 88.30 & 66.32 \\
\textbf{A2DINOv3 (Ours)}  &                         & \textbf{61.78} & \textbf{88.96} & \textbf{66.47} \\
\bottomrule
\end{tabular}
\caption{Object detection results on the M3FD dataset. Comparison of our proposed A2DINOv3 with baselines. All metrics are reported in percentage (\%). The best results within each modality group are highlighted in bold.}
\label{tab:main_m3fd1}
\end{table}
Table~\ref{tab:main_m3fd1} provides a detailed evaluation of our method specifically on the M3FD dataset, expanding the performance metrics to include different Intersection over Union (IoU) thresholds, namely mAP$_{50}$ and mAP$_{75}$. The results demonstrate that A2DINOv3 not only achieves the highest overall mAP (61.78\%) but also consistently outperforms both single-modality and multi-modality baselines under less strict (mAP$_{50}$) and more rigorous (mAP$_{75}$) localization criteria. Notably, the superior performance at mAP$_{75}$ (66.47\%) indicates that our method produces highly precise bounding box predictions. This validates the effectiveness of our regulated fusion approach, showing that selectively incorporating RGB cues with reliable infrared representations yields high-quality spatial localization alongside robust semantic recognition.

\subsection{GAIIC2024 Dataset Distribution Analysis}

To further illustrate the scale variations and domain shifts present in the GAIIC2024 dataset, we provide a visual and statistical comparison across the Training, Validation, and Test splits.

\begin{figure}[htbp]
    \centering
    \includegraphics[width=\columnwidth]{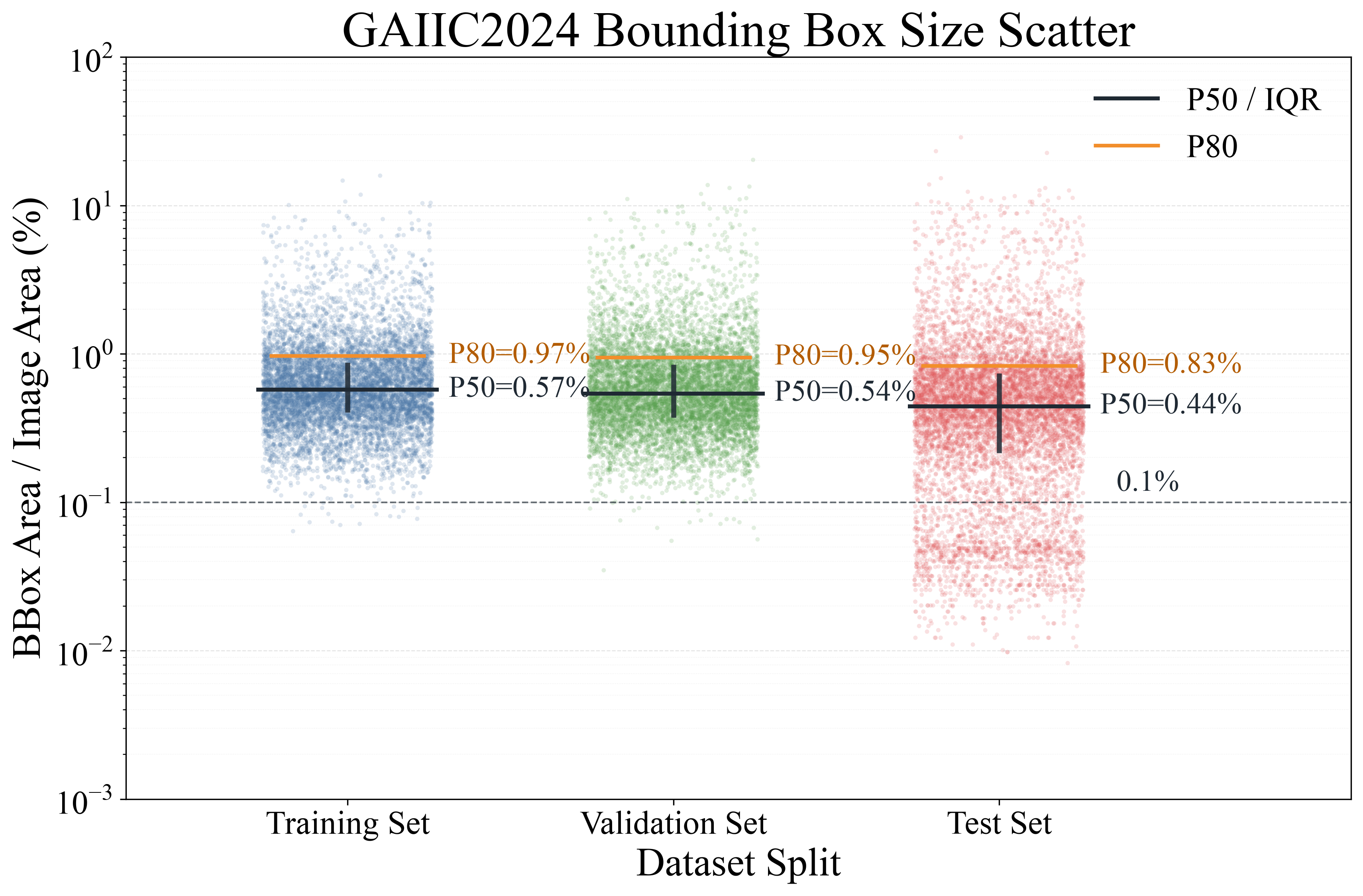}
    \caption{Scatter plot of relative bounding box areas across the dataset splits. }
    \label{fig:gaiic_scatter}
\end{figure}

\begin{figure*}[htbp]
    \centering
    \includegraphics[width=\textwidth]{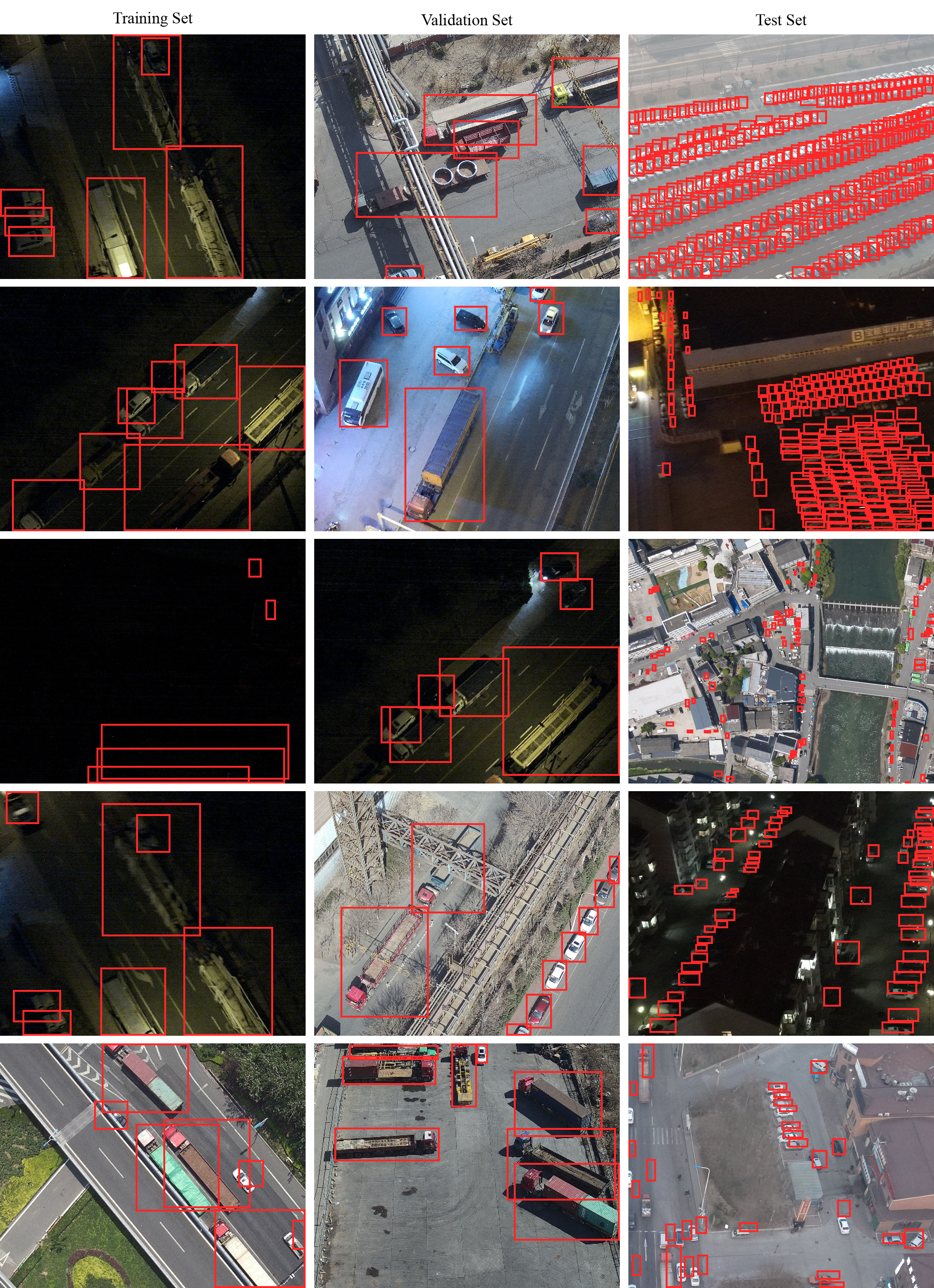}
    \caption{Visual comparison of the Training, Validation, and Test sets in the GAIIC2024 dataset. }
    \label{fig:gaiic_contrast}
\end{figure*}

\textbf{Quantitative Scale Variance.} The scatter plot in Fig~\ref{fig:gaiic_scatter} quantitatively confirms this observation by mapping the relative bounding box area (BBox Area / Image Area) for each split. 
\begin{itemize}
    \item Training and Validation Consistency: The Training and Validation sets share highly consistent object scale distributions, with median (P50) relative areas of 0.57\% and 0.54\%, respectively. Their 80th percentiles (P80) are also tightly aligned near 1.0\%.
    
    \item Test Set Degradation: The Test set exhibits a severe downward shift in object scale. The median relative area drops to 0.44\%, and the P80 drops to 0.83\%. 
    
\end{itemize}
\textbf{Visual Domain Shift.} As shown in Fig~\ref{fig:gaiic_contrast}, there is a distinct difference in the visual characteristics of the dataset splits. The training and validation sets typically feature closer-range perspectives, where target bounding boxes are relatively large and sparsely distributed across the frame. In stark contrast, the test set is dominated by high-altitude aerial perspectives, resulting in densely packed arrays of extremely small targets (e.g., tightly parked vehicles).


\bibliography{aaai2027}

\end{document}